\documentclass{ifacconf}

\pdfoutput=1 

\usepackage[utf8]{inputenc} 
\usepackage[T1]{fontenc}    
\usepackage{url}            
\usepackage{booktabs}       
\usepackage{amsfonts}       
\usepackage{amsmath}        
\usepackage{nicefrac}       
\usepackage{microtype}      
\usepackage{lipsum}
\usepackage{xcolor}
\usepackage{natbib}        
\usepackage{graphics} 
\usepackage{epsfig} 

\usepackage{tikz}
\usetikzlibrary{shapes.geometric, arrows}

\tikzstyle{block} = [rectangle, rounded corners, minimum width=1.5cm, minimum height=1cm,text centered, draw=white, fill=blue!30]

\tikzstyle{blockspecial} = [rectangle, rounded corners, minimum width=1.5cm, minimum height=1cm,text centered, draw=white, fill=orange!30]

\tikzstyle{operation} = [circle, rounded corners, minimum width=0.5cm, minimum height=0.5cm,text centered, draw=black, fill=white!30]

\tikzstyle{arrow} = [thick,->,>=stealth]
\begin{document}
\begin{frontmatter}

This work has been accepted by IFAC for publication (\textcopyright 2020 IFAC)

\title{Numerical Gaussian process Kalman filtering} 

\thanks[footnoteinfo]{The authors would like to thank the reviewers for their helpful comments.}

\author[First]{Armin K{\"u}per and Steffen Waldherr} 
\address[First]{Department of Chemical Engineering,
                KU Leuven, 
   Leuven, Belgium (e-mail: \{armin.kuper, steffen.waldherr\}@kuleuven.be).}

\begin{abstract}                
In this manuscript we introduce numerical Gaussian process Kalman filtering (GPKF).
Numerical Gaussian processes have recently been developed to simulate spatiotemporal models.
The contribution of this paper is to embed numerical Gaussian processes into the recursive Kalman filter equations.
This embedding enables us to do Kalman filtering on infinite-dimensional systems using Gaussian processes.
This is possible because i) we are obtaining a linear model from numerical Gaussian processes, and ii) the states of this model are by definition Gaussian distributed random variables.
Convenient properties of the numerical GPKF are that no spatial discretization of the model is necessary, and manual setting up of the Kalman filter, that is fine-tuning the process and measurement noise levels by hand is not required, as they are learned online from the data stream.
We showcase the capability of the numerical GPKF in a simulation study of the advection equation.
\end{abstract}


\end{frontmatter}

\section{Introduction}
Monitoring physical, chemical, or biological systems requires measurements obtained from sensors.
These measurements are subjected to stochastic noise.
In addition, technical or financial restrictions might prevent measuring all properties of interest.
State estimation methods help to overcome both of these obstacles.
They filter signal from noise, and reconstruct latent properties by leveraging mathematical models of the process at hand.
The most known state estimator is the Kalman filter, see \cite{kalman1960new}. 
It gives the optimal, in an expected squared error sense, state estimate for a linear system that is subjected to additive Gaussian noise.
State estimation methods for infinite-dimensional systems, that is systems described through partial differential equations (PDEs), are not that well established.
Usually, the PDEs are spatially discretized into large ODE systems that can be incorporated into the recursive Kalman filter equations.

In this work, we take a different route founded on Gaussian process (GP) regression, see \cite{williams2006gaussian}, and the recently introduced numerical GPs by \cite{raissi2018numerical}.
These are machine learning methods that unlike in a first principles derivation of a state space model, work with non-parametric models that adapt to the data.
A GP is a Gaussian distribution over functions and it is fully defined by a mean function and a covariance function.
When doing regression, a GP prior is placed on the regressor function.
The posterior distribution of the GP is then calculated by conditioning on measurements and estimating the so-called hyper-parameters of the covariance function.
An elegant extension to traditional GP regression to solve (non-)linear time-dependent PDEs, are numerical GPs.
In this method the PDE is first temporally discretized, hence the name numerical.
From this a multi-output GP is formulated whose structure, i.e. covariance functions, is informed by the time-discretized PDE.
Propagation through time is then simply done by iteratively forming the conditional distribution of this multi-output GP without having to do numerical integration.

Our contribution lies in embedding numerical GP regression into the recursive Kalman filter equations.
This is possible because the outputs of the numerical GP are, by definition, Gaussian distributed, and they relate linearly to each other through the time-discretized PDE.
We therefore obtain a Kalman filter for infinite-dimensional systems whose underlying process model is a first principles structured Gaussian process.
Tedious manual tuning of the Kalman filter is not necessary anymore as the process and measurement noise variance are learned dynamically on the data stream.

This paper is structured as follows: we start by introducing Kalman filtering and (numerical) GP regression in the methods section. 
Next, we show how numerical GP regression can be embedded into the recursive Kalman filter equations to obtain the numerical Gaussian process Kalman filter (GPKF).
Before giving a conclusion, we showcase the numerical GPKF using a simulation case study of the one-dimensional advection equation.

\subsubsection{Related works:}
Infinite-dimensional Kalman filtering via Gaussian process regression has been done previously by \cite{sarkka2012infinite} and \cite{sarkka2013spatiotemporal}.
Therein, spatiotemporal GP regression setups are converted into infinite-dimensional state space models.
These can then be used for infinite-dimensional Kalman filtering.
In numerical Gaussian process regression, the spatiotemporal model is first discretized in time and then formulated as a multi-output GP with the spatial coordinates as inputs.
Finite-dimensional filtering with Gaussian processes has been done, among others, by \cite{deisenroth2009analytic}, \cite{deisenroth2011general}, and \cite{ko2009gp}.

\section{Methods}
We give a brief introduction to Kalman filtering and (numerical) Gaussian process regression.
We use the respective common notations for Kalman filtering (state space models) and GP regression.
Both sections should therefore be read as separate entities.
The notation of numerical GP regression is oriented after the original work of \cite{raissi2018numerical} where the focus lies on spatiotemporal models.
In Section \ref{sec:numGPKF} we will unify both frameworks.

\subsection{Kalman filtering}
Kalman filtering can be used for stochastic filtering of a signal from noisy online measurements.
Furthermore, it can also be employed to reconstruct non-measurable states. 
It does so by using a probabilistic process and measurement model.
In case of linear process dynamics and an identical, independent Gaussian distributed noise acting on the process and measurements, the Kalman filter gives the optimal estimate of the states with respect to the expected squared error.

What follows is the Kalman filter from a Bayesian perspective as presented in \cite{sarkka2013bayesian}.

Suppose the process and measurement equations are given as a state space model
\begin{align}
  \boldsymbol{x}_t &= \boldsymbol{A} \boldsymbol{x}_{t-1} + \boldsymbol{q}_{t-1} \nonumber \\
  \boldsymbol{y}_t &= \boldsymbol{C} \boldsymbol{x}_t + \boldsymbol{r}_t.
  \label{eq:ssmodelKF}
\end{align}
Here, $\boldsymbol{x}_t \in \mathbb{R}^{d_x}$ is the state at time $t$ and $\boldsymbol{y}_t \in \mathbb{R}^{d_y}$ is the measurement at time $t$.
Process dynamics of the model are given by $\boldsymbol{A} \in \mathbb{R}^{d_x \times d_x}$ and the measurement model matrix is $\boldsymbol{C} \in \mathbb{R}^{d_y \times d_x}$.
Process noise $\boldsymbol{q}_{t-1} \in \mathbb{R}^{d_x}$ and measurement noise $\boldsymbol{r}_t \in \mathbb{R}^{d_y}$ are both modeled as white, additive Gaussian noise with diagonal covariance matrices $\boldsymbol{Q} \in \mathbb{R}^{d_x \times d_x}$ and $\boldsymbol{R} \in \mathbb{R}^{d_y \times d_y}$, respectively.
We formulate \eqref{eq:ssmodelKF} as probability density functions from which the Bayesian viewpoint of Kalman filtering naturally arises
\begin{align}
  p(\boldsymbol{x}_t | \boldsymbol{x}_{t-1}) & = \mathrm{N} (\boldsymbol{x}_t | \boldsymbol{A} \boldsymbol{x}_{t-1}, \boldsymbol{Q}), \nonumber \\
  p(\boldsymbol{y}_t|\boldsymbol{x}_{t}) & = \mathrm{N} (\boldsymbol{y}_t | \boldsymbol{C} \boldsymbol{x}_{t}, \boldsymbol{R}).
\end{align}
Here the first equation describes the stochastic dynamics of the system while the second equation gives the distribution of the current measurement given the current state.
These distributions are Gaussian for all times because the noise terms are Gaussian random variables, and the process and measurement equations are linear.

The goal in Bayesian filtering is to compute the marginal posterior distribution of the state $\boldsymbol{x}_t$ at each time step given all measurements up to the current time step.
The term marginal refers here to the marginalization over the previous state $\boldsymbol{x}_{t-1}$.
Using Bayes' theorem we have  
\begin{align}
  p(\boldsymbol{x}_t | \boldsymbol{y}_{1:t}) = \dfrac{p(\boldsymbol{y}_t | \boldsymbol{x}_t) \, p(\boldsymbol{x}_t | \boldsymbol{y}_{1:t-1})}{\int p(\boldsymbol{y}_t) | \boldsymbol{x}_t) \, p(\boldsymbol{x}_t | \boldsymbol{y}_{1:t-1}) \, \mathrm{d} \boldsymbol{x}_t}.
  \label{eq:recursivebayes}
\end{align}
With a constant data stream of measurements, this quickly becomes intractable as the measurement history grows ever larger.
The Kalman filter circumvents this problem by solving this equation recursively starting from a prior mean $\boldsymbol{m}_0$ and covariance $\boldsymbol{P}_0$.
The predictive, posterior, and normalizing distribution of \eqref{eq:recursivebayes} can be calculated in closed form. They are
\begin{align}
  p(\boldsymbol{x}_t | \boldsymbol{y}_{1:t-1}) &= \mathrm{N} (\boldsymbol{x}_t | \boldsymbol{m}_t^-, \boldsymbol{P}_t^-), \\
  p(\boldsymbol{x}_t | \boldsymbol{y}_{1:t}) &= \mathrm{N} (\boldsymbol{x}_t | \boldsymbol{m}_t, \boldsymbol{P}_t), \: \mathrm{and} \\
  p(\boldsymbol{y}_t | \boldsymbol{y}_{1:t-1}) &= \mathrm{N} (\boldsymbol{y}_t | \boldsymbol{C} \boldsymbol{m}_t^-, \boldsymbol{S}_t),
\end{align}
respectively. 
Moments of the above distributions are calculated in a prediction step
\begin{align}
\boldsymbol{m}_t^- &= \boldsymbol{A} \boldsymbol{m}_{t-1}, \\
\boldsymbol{P}_t^- &= \boldsymbol{A} \boldsymbol{P}_{t-1} \boldsymbol{A}^T + \boldsymbol{Q},
\end{align}
and an update step
\begin{align}
\boldsymbol{v}_t &= \boldsymbol{y}_t - \boldsymbol{C}\boldsymbol{m}_t^-,   &    \boldsymbol{m}_t &= \boldsymbol{m}_t^- + \boldsymbol{K}_t \boldsymbol{v}_t, \nonumber \\
\boldsymbol{S}_t &= \boldsymbol{C} \boldsymbol{P}_t^- \boldsymbol{C}^T + \boldsymbol{R},    &     \boldsymbol{P}_t &= \boldsymbol{P}^-_t - \boldsymbol{K}_t \boldsymbol{S}_t \boldsymbol{K}_t^T. \nonumber \\
\boldsymbol{K}_t &= \boldsymbol{P}_t^- \boldsymbol{C}^T \boldsymbol{S}^{-1}_t,    &    & 
\end{align}
One can derive these equations by formulating the required conditional, joint, and marginal distributions of the states and measurements.
This procedure can be found in \cite{sarkka2013bayesian} and we will use it later to derive the numerical Gaussian process Kalman filter equations.
\subsection{Gaussian process regression}
\label{subsec:gpr}
Here we give a brief introduction into Gaussian process regression where we take the function-space view presented in \cite{williams2006gaussian}. 
Afterwards we explain how numerical GP regression by \cite{raissi2018numerical} is used to solve time-dependent partial differential equations.

A GP is a Gaussian distribution over a random function $f(\boldsymbol{x})$.
It is fully defined through its mean function $m(\boldsymbol{x})$ and covariance function $k(\boldsymbol{x},\boldsymbol{x}')$
\begin{equation} \label{eq:gpdefinition}
\begin{array}{ll}
  m (\boldsymbol{x}) &= \mathrm{E} \left[ f(\boldsymbol{x}) \right] \\
  k(\boldsymbol{x},\boldsymbol{x}') &= \mathrm{E} \left[ \left( f(\boldsymbol{x}) - m(\boldsymbol{x}) \right) \left( f(\boldsymbol{x'}) - m(\boldsymbol{x'}) \right)^T \right].
\end{array} 
\end{equation}
Any finite dimensional collection of random variables $f(\boldsymbol{x}_1),\dotso,f(\boldsymbol{x}_l)$ is jointly Gaussian distributed
\begin{equation}
	\begin{pmatrix}
		f(\boldsymbol{x}_1) \\
		\vdots \\
		f(\boldsymbol{x}_l)
	\end{pmatrix}
	\sim \mathrm{N} \left(
	\begin{pmatrix}
		m(\boldsymbol{x}_1) \\
		\vdots \\
		m(\boldsymbol{x}_l)
	\end{pmatrix},
	\begin{pmatrix}
		k(\boldsymbol{x}_1,\boldsymbol{x}_1) & \cdots & k(\boldsymbol{x}_1,\boldsymbol{x}_l)  \\ 
		\vdots & \ddots & \vdots \\
		k(\boldsymbol{x}_l,\boldsymbol{x}_1) & \cdots & k(\boldsymbol{x}_l,\boldsymbol{x}_l) 
	\end{pmatrix}
	\right).
\end{equation}
Here, we introduce the notation $\boldsymbol{K} = \boldsymbol{K}(\boldsymbol{X},\boldsymbol{X})$, with $\boldsymbol{X}=(\boldsymbol{x}_1,\dotso,\boldsymbol{x}_l)$ and $\boldsymbol{K}_{ij} = k(\boldsymbol{x}_i,\boldsymbol{x}_j)$.
In regression we want to learn the function $f(\boldsymbol{x})$ from possibly noisy observations of its outputs at known inputs
\begin{equation}
y(\boldsymbol{x}_i) = f(\boldsymbol{x}_i) + \epsilon_i, \quad \epsilon_i \sim \mathrm{N}(0,\sigma^2_{\epsilon}).
\end{equation}
For GP regression we place a GP prior on $f(\boldsymbol{x})\sim \mathrm{GP}(m(\boldsymbol{x}),k(\boldsymbol{x},\boldsymbol{x'}))$ and formulate the conditional distribution of the function values to be predicted $\boldsymbol{f}(\boldsymbol{X}_*)=(f(\boldsymbol{x}_{*1}),\dotso, f(\boldsymbol{x}_{*p}))$ at test points $\boldsymbol{X}_*$, given the observations $\boldsymbol{y}(\boldsymbol{X})=(y(\boldsymbol{x}_1), \dotso, y(\boldsymbol{x}_l))$.
To obtain the conditional distribution, we first formulate the joint Gaussian distribution of the prediction and observation
\begin{equation}
	\label{eq:GPregressionjoint}
	\begin{pmatrix}
		\boldsymbol{f}(\boldsymbol{X}_*) \\
		\boldsymbol{y}(\boldsymbol{X})
	\end{pmatrix}
	\sim \mathrm{N} \left(
	\begin{pmatrix}
		\boldsymbol{m}(\boldsymbol{X}_*) \\
		\boldsymbol{m}(\boldsymbol{X})
	\end{pmatrix}
	,
	\begin{pmatrix}
		\boldsymbol{K}_{**} & \boldsymbol{K}_* \\
		\boldsymbol{K}_*^T & \boldsymbol{K}_y
	\end{pmatrix}
	\right),
\end{equation}
from which we get the conditional distribution using Lemma \ref{lem:conditional} (see Appendix)
\begin{align}
\label{eq:GPposterior}
\boldsymbol{f}(\boldsymbol{X}_*) | \boldsymbol{y}(\boldsymbol{X}) \sim \mathrm{N} ( & \boldsymbol{m}(\boldsymbol{X}_*) + \boldsymbol{K}_*\boldsymbol{K}_y^{-1} (\boldsymbol{y}(\boldsymbol{X})-\boldsymbol{m}(\boldsymbol{X})),  \nonumber \\
& \boldsymbol{K}_{**} - \boldsymbol{K}_*\boldsymbol{K}_y^{-1} \boldsymbol{K}_*^T ).
\end{align}
Here, measurement noise is accounted for inside $\boldsymbol{K}_y = \boldsymbol{K} + \sigma^2_{\epsilon} \boldsymbol{I}$. The covariance matrices in \eqref{eq:GPregressionjoint} and \eqref{eq:GPposterior} are $\boldsymbol{K}_{**}=\boldsymbol{K}(\boldsymbol{X}_*,\boldsymbol{X}_*) \in \mathbb{R}^{p \times p}$ and $\boldsymbol{K}_{*}=\boldsymbol{K}(\boldsymbol{X}_*,\boldsymbol{X}) \in \mathbb{R}^{p \times l}$.

Although GP regression is non-parametric, meaning that we do not need to define a structure for \( {f}(\boldsymbol{x}) \), we incorporate prior information through our choice of the kernel \( k(\boldsymbol{x},\boldsymbol{x}'; \boldsymbol{\theta} ) \) and its hyper-parameters \( \boldsymbol{\theta} \).
The squared exponential kernel (SE) finds broad application.
For scalar inputs it is
\begin{equation}
  k(x,x') = \sigma^2 \mathrm{exp}\left( \dfrac{(x-x')^2}{2l^2} \right).
\end{equation}
The hyper-parameters are the variance \( \sigma^2 \) and the length-scale \( l \).

The hyper-parameters $\boldsymbol{\theta}$ can be estimated from the measurement data by minimizing the negative log marginal likelihood function
\begin{equation}
\begin{array}{ll}
  -\mathrm{log}\,p(\boldsymbol{y} | \boldsymbol{X}) =& + \tfrac{1}{2} \boldsymbol{y}^T (\boldsymbol{K}(\boldsymbol{X},\boldsymbol{X};\boldsymbol{\theta}) + \sigma^2_{\epsilon}\boldsymbol{I})^{-1} \boldsymbol{y} \nonumber \\
  &+ \tfrac{1}{2}\mathrm{log}|\boldsymbol{K}(\boldsymbol{X},\boldsymbol{X};\boldsymbol{\theta}) + \sigma^2_{\epsilon}\boldsymbol{I}| + \tfrac{l}{2}\mathrm{log}\,2\pi.
\end{array}
\end{equation}
This provides a good compromise between data fit, first term, and model complexity, second term.

\subsubsection{Numerical Gaussian processes:}
In \cite{raissi2018numerical} numerical GPs have been introduced and used to solve spatiotemporal models.
Numerical GPs combine the data-driven machine learning nature of GPs with first principles knowledge from a spatiotemporal model.
Numerical refers to the fact that the spatiotemporal model has to be time-discretized.
For an easy to follow introduction of numerical GPs, we will be using the explicit Euler discretization here. 
For a general formulation, regardless of the discretization method, readers are referred to the original work of \cite{raissi2018numerical}.
To start, consider a linear partial differential equation
\begin{equation} \label{eq:pbe_linear}
  \dfrac{\partial n}{\partial t}(t,x) = \mathcal{L}_x n(t,x), 
\end{equation}
where \( \mathcal{L}_x \) is a linear operator acting on \( n(t,x) \) with respect to \( x \in \mathcal{R} \).
Discretization of \eqref{eq:pbe_linear} in time with the explicit Euler scheme yields
\begin{equation}
\begin{array}{ll}
  n_t &= n_{t-1} + \Delta t \mathcal{L}_x n_{t-1} \\
    &= \mathcal{Q}_x n_{t-1}.
    \label{eq:linpdedisc}
\end{array}
\end{equation}
We now place a GP prior of our choice on \( n_{t-1} \)
\begin{equation}
  n_{t-1} \sim \mathrm{GP} \left(0, k_{t-1,t-1}^{nn}(x,x') \right).
\end{equation}
It follows that
\begin{equation}
  n_t = \mathcal{Q}_x n_{t-1} \sim \mathrm{GP} \left(0,k_{t,t}^{nn} (x,x') \right),
\end{equation}
since a linear transformation of a GP is still a GP but with a different kernel that is informed by the linear transformation, see \cite{sarkka2011linear} for example.
Using the definition of a covariance function \eqref{eq:gpdefinition}, we can derive
\begin{equation}
\begin{array}{lll}
 k_{t,t} &= \mathrm{E} \left[ n_t(x) \left( n_t(x') \right)^T \right] \\
 &= \mathrm{E} \left[ \mathcal{Q}_x n_{t-1} (x) \left( \mathcal{Q}_{x'} n_{t-1} (x') \right)^T \right] \\
 &= \mathcal{Q}_x k_{t-1,t-1}^{nn} \mathcal{Q}_{x'}^T,
 \end{array}
\end{equation}
and 
\begin{equation}
\begin{array}{lll}
 k_{t,t-1}^{nn} &= \mathrm{E} \left[ n_t(x) \left( n_{t-1}(x') \right)^T \right] \\
 &= \mathrm{E} \left[ \mathcal{Q}_x n_{t-1} (x) \left(  n_{t-1} (x') \right)^T \right] \\
 &= \mathcal{Q}_x k_{t-1,t-1}^{nn}.
 \end{array}
\end{equation}
Note that if we place the prior on $n_{t}$, we would have to invert the linear operator.
Choosing where to place the GP prior is therefore crucial.

To perform temporal propagation, we first formulate the following multi-output GP
\begin{equation}\label{eq:multioutputGP}
  \begin{bmatrix}
    n_t \\
    n_{t-1}
  \end{bmatrix}
  \sim \mathrm{GP} \left(
  0,
  \begin{bmatrix}
    k_{t,t}^{nn} & k_{t,t-1}^{nn} \\
    k_{t-1,t}^{nn} & k_{t-1,t-1}^{nn}
  \end{bmatrix}
  \right).
\end{equation}
Starting from the initial condition, we now recursively calculate the conditional posterior distribution
\begin{equation}
p(n_t|n_{t-1}) = \mathrm{N} \left( \mu_t, \Sigma_{t,t} \right).
\end{equation}
The posterior mean and covariance are calculated as in \eqref{eq:GPposterior}.

Before formulating the conditional posterior distribution, the hyper-parameters have to be learned.
In the first time step, this is done on the initial and boundary data, and in the succeeding steps on artificially generated data, i.e. test points $\boldsymbol{X}_{t-1,*},\boldsymbol{X}_{t,*}$, and the current boundary data.
This artificial data has to be marginalized out from the posterior distribution to assure correct propagation of uncertainty.
The reader is referred to \cite{raissi2018numerical} for exact details of this procedure. 

Numerical GPs can handle explicit and implicit numerical methods. Merely \eqref{eq:multioutputGP} changes depending on the outputs and kernels introduced through the discretization method.
Furthermore, numerous types of boundary conditions and their combinations can be handled within this framework.

\section{Numerical Gaussian Process Kalman filtering}
\label{sec:numGPKF}
We start by extending numerical GPs to include online measurements and stochastic noise.
Then, we embed this structure into the traditional Kalman filter equations.

\subsection{Inclusion of a measurement equation and noise}
Consider a linear stochastic partial differential equation (PDE)
\begin{equation}
  \dfrac{\partial n}{\partial t} (t,x) = \mathcal{L}_x n(t,x) + q(t,x),
  \label{eq:pde}
\end{equation}
with $\mathcal{L}_x$ being a linear operator acting on the solution of the PDE $n(t,\cdot) \in \textit{L}_2 (\mathbb{R}^{d_x},\mathbb{R}),\, x \mapsto n(t,x)$, and $x \in \mathbb{R}^{d_x}$.
Process noise $q(t,x)$ is white, additive Gaussian noise.
To make the derivation easily accessible and without loss of generality, we use the explicit Euler time stepping scheme to derive the numerical GPKF.
Forward Euler discretization of \eqref{eq:pde} results in   
\begin{align}
 n_{t}(x) &= n_{t-1}(x) + \Delta t \mathcal{L}_x n_{t-1}(x) + \Delta t q_{t-1}(x), \nonumber \\
  &= \mathcal{F}_x n_{t-1}(x) + \Delta t q_{t-1}(x),
 \label{eq:temp_disc_pde}
\end{align}
where \( \mathcal{F}_x \) is again a linear operator summarizing the structure of the discretized PDE.
The right-hand side of \eqref{eq:pde} is not continuous, and one should therefore use the Euler-Maruyama method for discretization instead.
However, the covariance of the noise term $q_{t-1}$ will enter the algorithm by being multiplied with the step size $\Delta t$, much like in the Euler-Maruyama method, see $k^{nn}_{t,t}$ in Table~\ref{tab:kernels}.
We place mutually independent GP priors on $n_{t-1}$ and $q_{t-1}$
\begin{align}
  n_{t-1} &\sim \mathrm{GP} \left( 0, k_{t-1,t-1}^{nn}(x,x') \right), \\
  q_{t-1} &\sim \mathrm{GP} \left( 0, k_{t-1,t-1}^{qq}(x,x') \right),
\end{align}
with a kernel of our choice for $n_{t-1}$ and a white noise kernel $k_{t-1,t-1}^{qq}(x,x')=\delta (x-x') \sigma^2_q$ for $q_{t-1}$.
Other noise kernel choices are possible, but not sensible given the linear Gaussian setting.

We introduce a measurement equation
\begin{align}
  n_t^y(y) &= \mathcal{H}_x n_t(y) + r_t(y),
\end{align}
with the measurement operator $\mathcal{H}_x: \textit{L}_2 (\mathbb{R}^{d_x},\mathbb{R}) \rightarrow \textit{L}_2 (\mathbb{R}^{d_y},\mathbb{R}),\, n^t \mapsto n^t_y$, the measurement $y \in \mathbb{R}^{d_y}$, and additive, white noise $r_t(y)$ modeled as a GP with kernel $k_{t,t}^{rr}(y,y')=\delta (y-y') \sigma^2_r$.
Since $n_{t-1}$ is a GP, so is $n_t$, as already established, but so is also $n^y_t$.
Boundary conditions are usually a linear transformation of the state and are hence treated in the same fashion
\begin{equation}
  n_t^b(x_b) = \mathcal{B}_x n_t(x_b).
\end{equation}
We write the complete multi-output GP as
\begin{equation}
\begin{pmatrix}
  n_{t}\\
  n^y_t \\
  n_t^b \\
  n_{t-1}
\end{pmatrix}
\sim \mathrm{GP} \left(
\begin{pmatrix}
  0\\
  0\\
  0\\
  0
\end{pmatrix},
\begin{bmatrix}
k_{t,t}^{nn} & k_{t,t}^{n^y n} & k^{n n^b}_{t,t} & k_{t,t-1}^{nn} \\
k_{t,t}^{n^y n} & k_{t,t}^{n^y n^y} & k^{n^y n^b}_{t,t} & k_{t,t-1}^{n^y n} \\
k^{n^b n}_{t,t} & k^{n^b n^y}_{t,t} & k^{n^b n^b}_{t,t} & k^{n^b n}_{t,t-1} \\ 
k_{t-1,t}^{nn} & k_{t-1,t}^{nn^y} & k^{n n^b}_{t-1,t} & k_{t-1,t-1}^{nn} 
\end{bmatrix}
\right).
\label{eq:multioutputGPfull}
\end{equation}
These kernels, except for the boundary kernels, are listed in Table~\ref{tab:kernels}.
With this structure we are equipped to write down the recursive Kalman filter equations.
This is possible because we have a linear model whose states are Gaussian distributed random variables, i.e. Gaussian process distributed.

\begin{table}[tb]
\caption{Kernels of the multi-output explicit Euler GP \eqref{eq:multioutputGPfull}. For readability we write $k$ instead of $k_{t-1,t-1}^{nn}$.}
\begin{center}
  $
  \begin{array}{ll}
    \toprule
    \textbf{Kernel} & \textbf{Structure} \\
    \midrule
    k^{nn}_{t-1,t-1} & k \\
    k^{nn}_{t,t-1} & \mathcal{F}_x k \\
    k^{nn}_{t,t} & \mathcal{F}_x k \mathcal{F}_{x'}^T + \Delta t^2 k_{t-1,t-1}^{qq} \\
    k^{n^y n}_{t,t-1} & \mathcal{H}_x \mathcal{F}_x k \\
    k^{n n^y}_{t,t} & \mathcal{F}_x k \left( \mathcal{F}_{x'} \mathcal{H}_{x'} \right)^T + \Delta t^2 k_{t-1,t-1}^{qq} \mathcal{H}_{x'}^T \\
    k^{n^y n^y}_{t,t} & 
    \mathcal{H}_x \mathcal{F}_x k \left( \mathcal{F}_{x'} \mathcal{H}_{x'} \right)^T + \Delta t^2 \mathcal{H}_x k_{t-1,t-1}^{qq} \mathcal{H}_{x'}^T + k_{t,t}^{rr} \\
    \bottomrule
  \end{array}$
  \label{tab:kernels}
  \end{center}
\end{table}

\subsection{The numerical Gaussian process Kalman filter}
In this subsection we only work with finite-dimensional collections of the Gaussian process random variables. 
We therefore write $\mathrm{N}(\cdot,\cdot)$ instead of $\mathrm{GP}(\cdot,\cdot)$.
Furthermore, we use bold lowercase letters for the random variables and kernels become covariance matrices, indicated by bold capital letters, as already introduced in subsection \ref{subsec:gpr}.

\begin{figure}[tb]
  \begin{center}
  \resizebox{8.4cm}{!}{%
    \begin{tikzpicture}[node distance=2cm]
      \node (priormodel) [block] {$p(n_t|n_{t-1})$};
      \node (times1) [operation, right of=priormodel] {$\times$};
      \node (previouspost) [block, right of=times1] {$p(n_{t-1}|n^y_{1:t-1})$};      
      \node (priorjoint) [block, below of=times1] {$p(n_t,n_{t-1}|n^y_{1:t-1})$};
      \node (prior) [blockspecial, below of=priorjoint] {\textbf{prior} $p(n_t | n^y_{1:t-1})$};
      \node (times2) [operation, right of=prior, xshift=5cm] {$\times$};
      \node (likelihood) [blockspecial, below of=times2, yshift=0.5cm] {\textbf{likelihood} $p(n^y_t|n_t)$};
      \node (posteriorjoint) [block, above of=times2] {$p(n_t, n^y_t | n^y_{1:t-1})$};
      \node (posterior) [blockspecial, above of =posteriorjoint] {\textbf{posterior} $p(n_t|n^y_{1:t})$};

      \draw [arrow] (priormodel) -- (times1);
      \draw [arrow] (previouspost) -- (times1);
      \draw [arrow] (times1) -- (priorjoint);
      \draw [arrow] (priorjoint) -- node[anchor=east] {marginalizing} (prior);
      \draw [arrow] (prior) -- (times2);
      \draw [arrow] (likelihood) -- (times2);
      \draw [arrow] (times2) -- (posteriorjoint);    
      \draw [arrow] (posteriorjoint) -- node[anchor=east] {conditioning} (posterior);  
      \draw [arrow] (posterior) -- node[anchor=south, text width=1.5cm] {repeat recursively} (previouspost);        
    \end{tikzpicture}
    }
    \caption{Recursive posterior distribution calculation. Boundary data has been omitted for readability as it is treated in the same fashion as measurements.}
    \label{fig:recursivefilter}
  \end{center}
\end{figure}
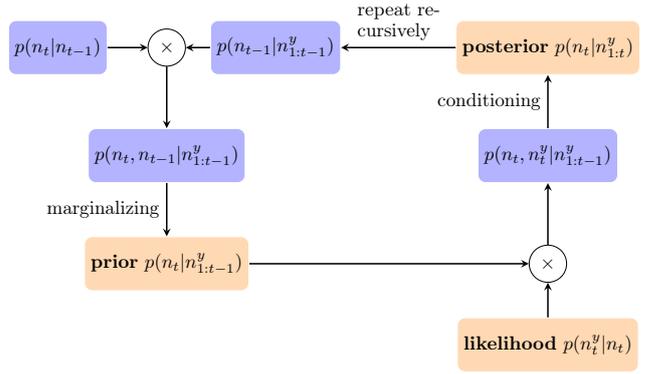

Recall that in Kalman filtering we want to calculate the posterior distribution of a dynamic state given model predictions and measurements up to the current time.
According to Bayes' rule the posterior is given by
\begin{equation}
p(\boldsymbol{n}_t|\boldsymbol{n}^y_{1:t}) = \dfrac{p(\boldsymbol{n}^y_t|\boldsymbol{n}_t) p(\boldsymbol{n}_t|\boldsymbol{n}_{1:t-1}^y)}{p(\boldsymbol{n}^y_t|\boldsymbol{n}^y_{1:t-1})}.
\end{equation}
Here we omitted the boundary data $\boldsymbol{n}^b_t(x_b)$ for readability.
We assume the states to be Markovian, i.e. the current state $\boldsymbol{n}_t$ is conditionally independent of anything that happened before $t-1$.
Furthermore, the current measurement given the current state is conditionally independent of the measurement and state histories.
The individual terms can be calculated in closed form
\begin{itemize}
  \item prior \( p(\boldsymbol{n}_t|\boldsymbol{n}_{t-1}^y) = \mathrm{N}(\boldsymbol{n}_t|\boldsymbol{m}^-_t, \boldsymbol{P}^-_t) \),
  \item posterior \( p(\boldsymbol{n}_t|\boldsymbol{n}^y_t) = \mathrm{N}(\boldsymbol{n}_t|\boldsymbol{m}_t, \boldsymbol{P}_t) \),
  \item normalizing constant \( p(\boldsymbol{n}^y_t|\boldsymbol{n}^y_{t-1}) = \mathrm{N}(\boldsymbol{n}^y_t|\mathcal{H}_x \boldsymbol{m}^-_t, \boldsymbol{P}_t) \).
\end{itemize}
An illustration of the road map ahead is shown in Fig.~\ref{fig:recursivefilter}.

To calculate the prior distribution, we first calculate the joint distribution of the states $\boldsymbol{n}_{t-1}$ and $\boldsymbol{n}_t$ conditioned on the measurement and boundary data history
\begin{align}
  p(\boldsymbol{n}_{t-1},\boldsymbol{n}_t|\boldsymbol{n}^y_{1:t-1}, \boldsymbol{n}^b_{1:t-1}) &= \underbrace{p(\boldsymbol{n}_t|\boldsymbol{n}_{t-1})}_{\text{model prediction}} \nonumber \\
   & \quad \times \underbrace{p(\boldsymbol{n}_{t-1}|\boldsymbol{n}^y_{1:t-1}, \boldsymbol{n}^b_{1:t-1})}_{\text{previous posterior}} \nonumber \\
  &= \mathrm{N} (\boldsymbol{n}_t | \boldsymbol{A} \boldsymbol{n}_{t-1}, \boldsymbol{P}^{GP,nn}_t) \nonumber \\
  &\quad \times \mathrm{N}\left( \boldsymbol{n}_{t-1}| \boldsymbol{m}_{t-1}, \boldsymbol{P}_{t-1} \right) \nonumber \\
  & = \mathrm{N} \left( \begin{pmatrix}
  \boldsymbol{n}_{t-1} \\
  \boldsymbol{n}_t
  \end{pmatrix} | \boldsymbol{m}', \boldsymbol{P}' \right). 
  \label{eq:priorjoint}
\end{align}
Using Lemma \ref{lem:jointgaussian}, the joint mean of \eqref{eq:priorjoint} is
\begin{equation}
  \boldsymbol{m}' = \begin{pmatrix}
  \boldsymbol{m}_{t-1} \\
  \boldsymbol{A} \boldsymbol{m}_{t-1}
  \end{pmatrix}
\end{equation}
and the covariance is
\begin{equation}
  \boldsymbol{P}' = \begin{bmatrix}
    \boldsymbol{P}_{t-1} & \boldsymbol{P}_{t-1} \boldsymbol{A}^T \\
    \boldsymbol{A} \boldsymbol{P}_{t-1} & \boldsymbol{A} \boldsymbol{P}_{t-1} \boldsymbol{A}^T + \boldsymbol{P}^{GP,nn}_t
  \end{bmatrix}.
\end{equation}
We introduced
\begin{equation}
  \boldsymbol{A} = \boldsymbol{K}_{t,t-1}^{nn} (\boldsymbol{K}_{t-1,t-1}^{nn})^{-1},
  \label{eq:pseudodyn}
\end{equation}
and
\begin{equation}
  \boldsymbol{P}^{GP,nn}_t = \boldsymbol{K}_{t,t}^{nn} - \boldsymbol{K}^{nn}_{t,t-1} (\boldsymbol{K}_{t-1,t-1}^{nn})^{-1} \boldsymbol{K}^{nn}_{t-1,t}.
\end{equation}
These two equations are obtained by forming the conditional $\boldsymbol{n}_t|\boldsymbol{n}_{t-1}$ of the multi-output GP \eqref{eq:multioutputGP}.
We can think of \eqref{eq:pseudodyn} as the equivalent to the dynamic matrix in a state space model with the difference that its entries depend on the test points $A(\boldsymbol{X}_{t,*},\boldsymbol{X}_{t-1,*})$.
Now, the prior distribution is obtained by marginalizing over $n_{t-1}$ in \eqref{eq:priorjoint}
\begin{equation}
  p(\boldsymbol{n}_t| \boldsymbol{n}^y_{1:t-1}, \boldsymbol{n}^b_{1:t-1}) = \mathrm{N} \left( \boldsymbol{n}_t | \boldsymbol{m}^-_t, \boldsymbol{P}^-_t \right),
\end{equation}
with prior mean and covariance
\begin{align}
  \boldsymbol{m}^-_t &= \boldsymbol{A} \boldsymbol{m}_{t-1}, \\
  \boldsymbol{P}^-_t &= \boldsymbol{A} \boldsymbol{P}_{t-1} \boldsymbol{A}^T + \boldsymbol{P}^{GP,nn}_t.
\end{align}
Equipped with this, we can calculate the joint distribution $p(\boldsymbol{n}_t,\boldsymbol{n}_t^y,\boldsymbol{n}_t^b|\boldsymbol{n}^y_{1:t-1},\boldsymbol{n}^b_{1:t-1})$.
This will allow us to write down the posterior distribution later.
We start by noting that
\begin{align}
  p(\boldsymbol{n}_t, \boldsymbol{n}^y_t, \boldsymbol{n}^b_t|\boldsymbol{n}^y_{1:t-1}, \boldsymbol{n}^b_{1:t-1}) &= p(\boldsymbol{n}^y_t, \boldsymbol{n}^b_t|\boldsymbol{n}_t) \nonumber \\
  &\quad \times p(\boldsymbol{n}_t | \boldsymbol{n}^y_{1:t-1}, \boldsymbol{n}^b_{1:t-1}) \nonumber \\
  &= \mathrm{N} ( \boldsymbol{n}^y_t, \boldsymbol{n}^b_t | \boldsymbol{C} \boldsymbol{n}_t, \boldsymbol{P}^{GP,n^y n^y}_t ) \nonumber \\
  &\quad \times \mathrm{N}\left(\boldsymbol{n}_t | \boldsymbol{m}^-_t, \boldsymbol{P}^-_t \right) \nonumber \\
  &= \mathrm{N} \left( \begin{pmatrix}
  \boldsymbol{n}_t \\
  \boldsymbol{n}^y_t \\
  \boldsymbol{n}^b_t \end{pmatrix} | \boldsymbol{m}'', \boldsymbol{P}'' \right).
\end{align}
Here we introduced
\begin{align}
\label{eq:pseudomeasmatrix}
\boldsymbol{C} = 
\begin{bmatrix}
  \boldsymbol{K}^{n^y n}_{t,t}\\
  \boldsymbol{K}^{n^b n}_{t,t}
  \end{bmatrix} (\boldsymbol{K}^{nn}_{t,t})^{-1},
\end{align}
and 
\begin{align}
\boldsymbol{P}^{GP,n^y n^y}_t =&
\begin{bmatrix}
  \boldsymbol{K}^{n^y n^y}_{t,t} & \boldsymbol{K}^{n^y n^b}_{t,t} \\
  \boldsymbol{K}^{n^b n^y}_{t,t} & \boldsymbol{K}^{n^b n^b}_{t,t}
  \end{bmatrix} \nonumber \\
  &- \begin{bmatrix}
  \boldsymbol{K}^{n^y n}_{t,t} \\
  \boldsymbol{K}^{n^b n}_{t,t}
  \end{bmatrix} (\boldsymbol{K}^{nn}_{t,t})^{-1} \begin{bmatrix}
  \boldsymbol{K}^{n n^y}_{t,t} & \boldsymbol{K}^{n n^b}_{t,t}
  \end{bmatrix}.
\end{align}
These two expressions are obtained in the similar fashion as $\boldsymbol{A}$ and $\boldsymbol{P}_t^{GP,nn}$ by forming $\boldsymbol{n}^y_t,\boldsymbol{n}^b_t|\boldsymbol{n}_t$ from \eqref{eq:multioutputGPfull}.
Again, we can think of $\boldsymbol{C}$ as the measurement matrix equivalent in a state space model, albeit its entries depend on the measurement, boundary, and test point locations $\boldsymbol{C}(\boldsymbol{Y}_t,\boldsymbol{X}_{t,b},\boldsymbol{X}_{t,*})$.
Using Lemma \ref{lem:jointgaussian} one more time, the joint mean is
\begin{equation}
  \boldsymbol{m}'' = \begin{pmatrix}
    \boldsymbol{m}^-_t \\
    \boldsymbol{C} \boldsymbol{m}^-_t
    \end{pmatrix}.
\end{equation}
The covariance is
\begin{equation}
\boldsymbol{P}'' = \begin{bmatrix}
  \boldsymbol{P}^-_t & \boldsymbol{P}^-_t \boldsymbol{C}^T \\
  \boldsymbol{C} \boldsymbol{P}^-_t & \boldsymbol{C} \boldsymbol{P}^-_t \boldsymbol{C}^T + \boldsymbol{P}^{GP,n^y n^y}_t.
\end{bmatrix}
\end{equation}
To get the posterior $p(\boldsymbol{n}_t|\boldsymbol{n}^y_{1:t},\boldsymbol{n}^b_{1:t})$, we need to condition the above joint distribution $p(\boldsymbol{n}_t, \boldsymbol{n}^y_t, \boldsymbol{n}^b_t|\boldsymbol{n}^y_{1:t-1}, \boldsymbol{n}^b_{1:t-1})$ on the current measurement and boundary data using Lemma \ref{lem:conditional}
\begin{align}
p(\boldsymbol{n}_t | \boldsymbol{n}^y_t, \boldsymbol{n}^b_t, \boldsymbol{n}^y_{1:t-1}, \boldsymbol{n}^b_{1:t-1}) &= p(\boldsymbol{n}_t | \boldsymbol{n}^y_{1:t}, \boldsymbol{n}^b_{1:t}) \nonumber \\
  &= \mathrm{N} \left( \boldsymbol{m}_t, \boldsymbol{P}_t \right).
\end{align}
The posterior mean is \footnote{Careful readers will recognize the structure of this posterior KF estimate from the posterior GP mean and covariance, compare with \eqref{eq:GPposterior}. This shouldn't be a surprise, as in both cases a conditional Gaussian is formed.}
\begin{align}
 \boldsymbol{m}_t &= \boldsymbol{m}^-_t  + \boldsymbol{P}^-_t \boldsymbol{C}^T \nonumber \\
  & \quad \times \left( \boldsymbol{C} \boldsymbol{P}^-_t \boldsymbol{C}^T  + \boldsymbol{P}^{GP,n^y n^y}_t \right)^{-1} \left( [\boldsymbol{n}^y_t \; \boldsymbol{n}^b_t]^T - \boldsymbol{C} \boldsymbol{m}^-_t \right),
\end{align}
and the corresponding variance is 
\begin{equation}
  \boldsymbol{P}_t = \boldsymbol{P}^-_t - \boldsymbol{P}^-_t \boldsymbol{C}^T \left( \boldsymbol{C} \boldsymbol{P}^-_t \boldsymbol{C}^T + \boldsymbol{P}^{GP,n^y n^y}_t \right)^{-1} \boldsymbol{C} \boldsymbol{P}^-_t.
\end{equation}
To summarize the numerical GPKF, we have the prediction step as
\begin{align}
  \boldsymbol{m}^-_t &= \boldsymbol{A} \boldsymbol{m}^{-}_{t-1}, \\
  \boldsymbol{P}^-_t &= \boldsymbol{A} \boldsymbol{P}_{t-1} \boldsymbol{A}^T + \boldsymbol{P}^{GP,nn}_t,
\end{align}
and the update step as
\begin{align}
\boldsymbol{v}_t &= \boldsymbol{n}^y_t - \boldsymbol{C} \boldsymbol{m}^-_t, \\
\boldsymbol{S}_t &= \boldsymbol{C} \boldsymbol{P}_t^- \boldsymbol{C}^T + \boldsymbol{P}^{GP,n^y n^y}_t, \\
\boldsymbol{K}_t &= \boldsymbol{P}^-_t \boldsymbol{C}^T (\boldsymbol{S}_t)^{-1}, \\
\boldsymbol{m}_t &= \boldsymbol{m}^-_t + \boldsymbol{K}_t \boldsymbol{v}_t, \\
\boldsymbol{P}_t &= \boldsymbol{P}_t^- - \boldsymbol{K}_t \boldsymbol{S}_t (\boldsymbol{K}_t)^T.
\end{align}
The covariances of process and measurement noise are added to the kernels $\boldsymbol{K}^{nn}_{t,t}$ and $\boldsymbol{K}^{n^y n^y}_{t,t}$, as shown in Table \ref{tab:kernels}.
They therefore enter the algorithm as in the traditional Kalman filter.

By minimizing the negative log-marginal likelihood, hyper-parameters are estimated in every time step from the current measurement, boundary, and previous test data. 

\section{Simulation study}
\begin{figure}[tb]
  \begin{center}
  \includegraphics[width=8.4cm]{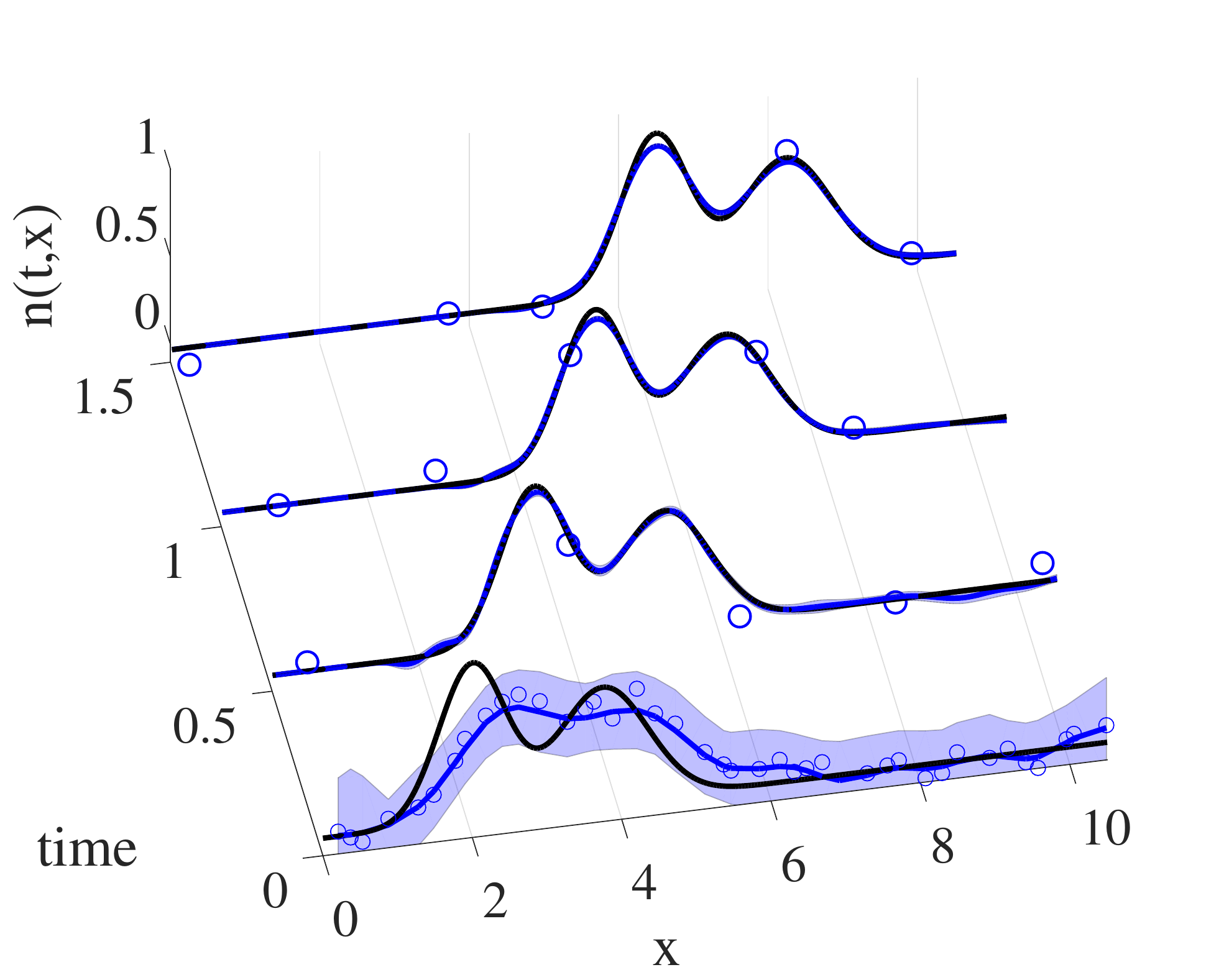}    
  \caption{Temporal snapshots of the analytical solution (black) and the posterior estimate of the numerical GPKF (blue) along with two times its standard deviation. Blue circles indicate measurements. $41$ training points are sampled from the initial estimate, from then on $5$ noisy measurements are available in each time step $\Delta t = 5 \cdot 10^{-3}$.}
  \label{fig:snapshots_1D}
  \end{center}
\end{figure}
We will look at the one-dimensional advection equation
\begin{equation}
  \dfrac{\partial n}{\partial t}(t,x) + gn(t,x) = 0. 
  \label{eq:pbecasestudy}
\end{equation}
Here, the advection speed is $g=3$.
We impose a no-flux boundary condition $n(t,x_b=0) g=0$.
The initial state is a bimodal Gaussian distribution $n(0,x) = \mathrm{N}(\mu_{0,1}=2,\sigma^2_{0,1}=0.45^2) + \mathrm{N}(\mu_{0,2} = 3.75, \sigma^2_{0,2} = 0.6^2)$.

For this numerical case study, we assume to obtain noise corrupted point evaluations of the analytical solution
\begin{equation}
  n^y_t(y_i) = n_t(y_i) + \epsilon_i. 
\end{equation} 
Here $\epsilon$ is generated from a zero mean Gaussian with variance $\sigma^2_{\epsilon}=0.06^2$.
Process noise is zero.
\begin{figure}[tb]
  \begin{center}
  \includegraphics[width=8.4cm]{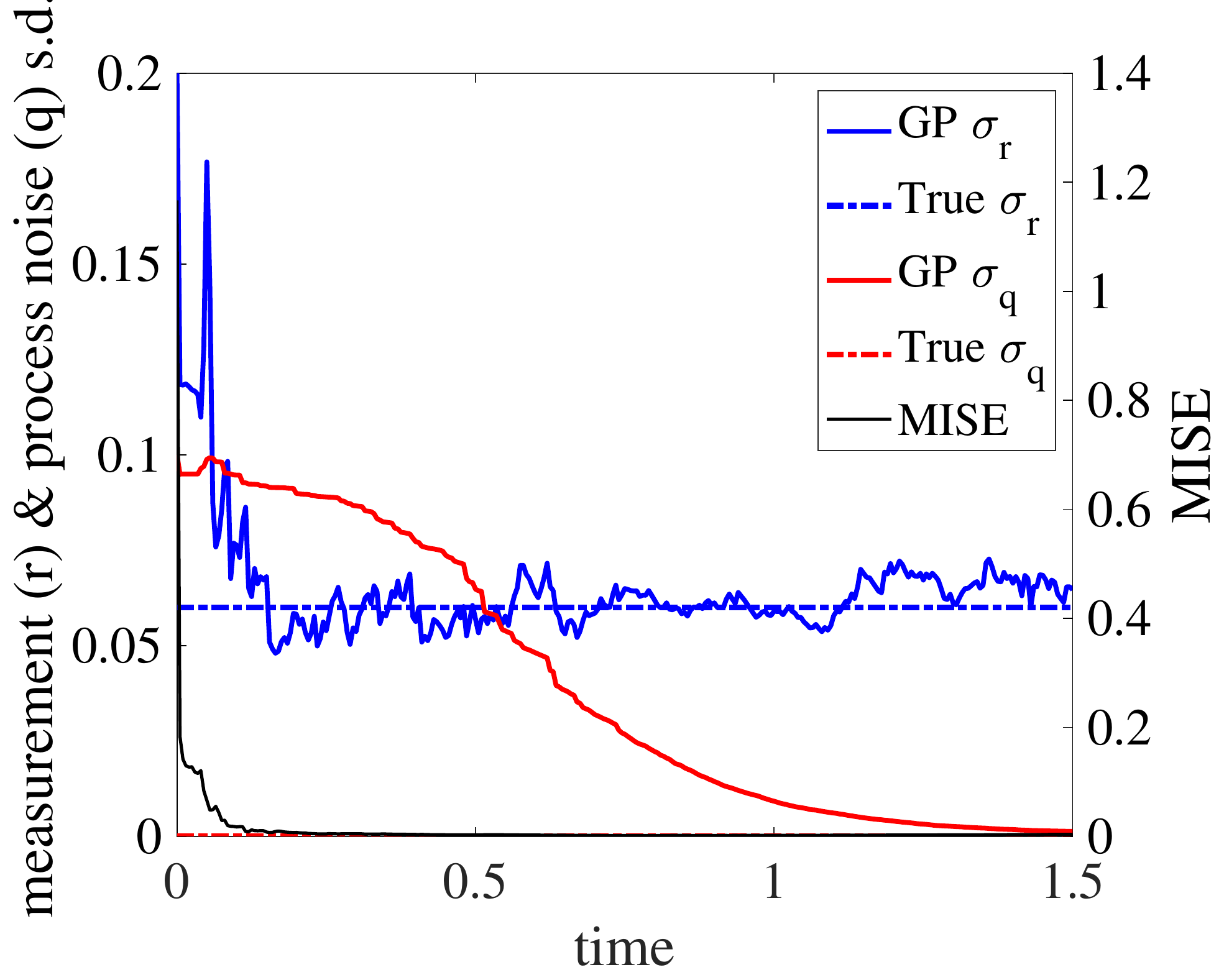}    
  \caption{Process and measurement noise standard deviation estimation, as well as the mean integrated squared error between posterior state estimation and ground truth.}
  \label{fig:hyper_mise}
  \end{center}
\end{figure}

We discretize \eqref{eq:pbecasestudy} using the implicit Euler method with a step size of $\Delta t=5 \cdot 10^{-3}$.
This is also used as the sampling time for measurements.
A GP prior with a zero-mean function and squared exponential kernel is placed on $n_t(x)$.
From this, the entire structure of the numerical GPKF is built up.

The initial state estimate is a shifted and diffused version of the true initial state $n(0,x)=\mathrm{N}(\boldsymbol{\mu}_0+0.5,\boldsymbol{\sigma}^2_0+0.2)$.
From the initial estimate, $N_{tr}=41$ points are sampled and regular GP regression is performed.
This gives the initial variance $\boldsymbol{P}_0$ of the Kalman filter.
Initial hyper-parameter values of the squared exponential are $l_{SE}=0.5$ and $\sigma^2_{SE}=0.3^2$, and $\sigma^2_{q}=0.1^2$ for the process white noise kernel, and $\sigma^2_{r}=0.2^2$ for the measurement white noise kernel.

Fig.~\ref{fig:snapshots_1D} shows temporal snapshots of the analytical solution and the numerical GPKF posterior estimate along with two times its standard deviation. 
State estimation convergence towards ground truth in terms of the mean integrated squared error (MISE) can be seen in Fig.~\ref{fig:hyper_mise}. 
Therein, the adaptation of the noise hyper-parameters is shown as well. 
\section{Conclusion}
In this paper, we introduced numerical Gaussian process Kalman filtering (GPKF) with which we can do state and noise estimation for infinite-dimensional systems.
Numerical Gaussian processes by \cite{raissi2018numerical} solve spatiotemporal models with Gaussian process regression by discretization in time and leveraging the resulting structure for kernel design.
We embedded this method into the well known recursive Kalman filter equations.

Learning the noise hyper-parameters online on the data stream could be a two-sided sword.
In cases where the negative log-marginal likelihood function has a rugged contour, performance of the numerical GPKF could be drastically reduced by getting stuck in local minima.
Further simulation studies with more complex dynamics should investigate this, as well as non-zero process noise.

                                                                   



\appendix
%
\section{Gaussian identities}
The following is taken from \cite{sarkka2013bayesian}.
\begin{lem}
\label{lem:jointgaussian}
(Joint distribution of Gaussian variables) If random variables $\boldsymbol{x} \in \mathbb{R}^{d_x}$ and $\boldsymbol{y} \in \mathbb{R}^{d_y}$ have the Gaussian probability distributions
\begin{align}
  \boldsymbol{x} &\sim \mathrm{N} \left(\boldsymbol{m}, \boldsymbol{P} \right) \\
  \boldsymbol{y|x} &\sim \mathrm{N} \left(\boldsymbol{H}\boldsymbol{x} + \boldsymbol{u}, \boldsymbol{R} \right),
\end{align}
than the joint distribution of $\boldsymbol{x},\,\boldsymbol{y}$ and the marginal distribution of $\boldsymbol{y}$ are given as
\begin{align}
  \begin{pmatrix}
    \boldsymbol{x} \\
    \boldsymbol{y}
    \end{pmatrix} &\sim \mathrm{N} \left(
    \begin{pmatrix}
      \boldsymbol{m} \\
      \boldsymbol{H} \boldsymbol{m} + \boldsymbol{u}
    \end{pmatrix}, 
    \begin{pmatrix}
      \boldsymbol{P} & \boldsymbol{P} \boldsymbol{H}^T \\
      \boldsymbol{H}\boldsymbol{P} & \boldsymbol{H}\boldsymbol{P}\boldsymbol{H}^T + \boldsymbol{R}
    \end{pmatrix}
    \right), \\
    \boldsymbol{y} &\sim \mathrm{N} \left( \boldsymbol{H}\boldsymbol{m} + \boldsymbol{u}, \boldsymbol{H}\boldsymbol{P}\boldsymbol{H}^T + \boldsymbol{R} \right).
\end{align}
\end{lem}
\begin{lem}
\label{lem:conditional}
(Conditional distribution of Gaussian variables) If the random variables $\boldsymbol{x}$ and $\boldsymbol{y}$ have the joint Gaussian probability distribution
\begin{align}
  \begin{pmatrix}
    \boldsymbol{x} \\
    \boldsymbol{y}
    \end{pmatrix} &\sim \mathrm{N} \left(
    \begin{pmatrix}
      \boldsymbol{a} \\
      \boldsymbol{b}
    \end{pmatrix}, 
    \begin{pmatrix}
      \boldsymbol{A} & \boldsymbol{C} \\
      \boldsymbol{C}^T & \boldsymbol{B}
    \end{pmatrix}
    \right),
\end{align}
than the conditional distribution is
\begin{equation}
  \boldsymbol{x}|\boldsymbol{y} \sim \mathrm{N} \left( \boldsymbol{a} + \boldsymbol{C}\boldsymbol{B}^{-1}(\boldsymbol{y}-\boldsymbol{b}), \boldsymbol{A} - \boldsymbol{C}\boldsymbol{B}^{-1}\boldsymbol{C}^T \right).
\end{equation}
\end{lem}
\end{document}